\documentclass{article}

\usepackage{microtype}

\usepackage{graphicx}
\usepackage{subcaption}
\usepackage{booktabs} 

\PassOptionsToPackage{hyphens}{url}
\PassOptionsToPackage{breaklinks=true}{hyperref}
\usepackage{hyperref}

\usepackage[preprint]{icml2026}

\usepackage{amsmath}
\usepackage{amssymb}
\usepackage{mathtools}
\usepackage{amsthm}

\usepackage[capitalize,noabbrev]{cleveref}

\usepackage{algorithm}
\usepackage{algorithmic}
\usepackage{listings}
\usepackage{xcolor}
\usepackage{tikz}
\usetikzlibrary{positioning, arrows.meta, calc}

\lstdefinestyle{prompt}{
  basicstyle=\footnotesize\ttfamily,
  backgroundcolor=\color[gray]{0.96},
  frame=single,
  rulecolor=\color[gray]{0.75},
  framerule=0.4pt,
  breaklines=true,
  breakatwhitespace=true,
  breakindent=0pt,
  showstringspaces=false,
  keepspaces=true,
  columns=flexible,
  xleftmargin=0.6em,
  xrightmargin=0.6em,
  aboveskip=0.6em,
  belowskip=0.6em,
  literate=%
    {—}{{---}}1 {–}{{--}}1
    {ℝ}{{$\mathbb{R}$}}1 {ℤ}{{$\mathbb{Z}$}}1 {ℕ}{{$\mathbb{N}$}}1 {ℚ}{{$\mathbb{Q}$}}1 {ℂ}{{$\mathbb{C}$}}1
    {≥}{{$\geq$}}1 {≤}{{$\leq$}}1 {≠}{{$\neq$}}1
    {∈}{{$\in$}}1 {∉}{{$\notin$}}1 {∀}{{$\forall$}}1 {∃}{{$\exists$}}1
    {→}{{$\to$}}1 {↦}{{$\mapsto$}}1 {⊆}{{$\subseteq$}}1 {⊂}{{$\subset$}}1
    {…}{{\ldots}}1 {π}{{$\pi$}}1 {θ}{{$\theta$}}1 {α}{{$\alpha$}}1 {β}{{$\beta$}}1 {ε}{{$\varepsilon$}}1
    {⌊}{{$\lfloor$}}1 {⌋}{{$\rfloor$}}1 {⟨}{{$\langle$}}1 {⟩}{{$\rangle$}}1
    {·}{{$\cdot$}}1 {×}{{$\times$}}1 {∧}{{$\wedge$}}1 {∨}{{$\vee$}}1 {¬}{{$\neg$}}1
    {“}{{``}}1 {”}{{''}}1 {‘}{{`}}1 {’}{{'}}1
}

\theoremstyle{plain}

\theoremstyle{definition}

\theoremstyle{remark}

\usepackage[textsize=tiny]{todonotes}

\icmltitlerunning{Goedel-Architect}

\begin{document}

\twocolumn[
  \icmltitle{Goedel-Architect: Streamlining Formal Theorem Proving with\\Blueprint Generation and Refinement}



  \icmlsetsymbol{equal}{*}
  \icmlsetsymbol{last}{\dag}

  \begin{icmlauthorlist}
    \icmlauthor{Jui-Hui Chung}{equal,pli}
    \icmlauthor{Ziyang Cai}{equal,pli}
    \icmlauthor{Zihao Li}{pli}
    \icmlauthor{Qishuo Yin}{pli}
    \icmlauthor{Rohit Agarwal}{pli}
    \icmlauthor{Simon Park}{pli}
    \icmlauthor{Rodrigo Porto}{pli}
    \icmlauthor{Narutatsu Ri}{pli}
    \icmlauthor{Ziran Yang}{pli}
    \icmlauthor{Shange Tang}{pli}
    \icmlauthor{Xingyu Dang}{pli}
    \icmlauthor{Hongzhou Lin}{amazon}
    \icmlauthor{Mengdi Wang}{pli}
    \icmlauthor{Danqi Chen}{pli}
    \icmlauthor{Chi Jin}{pli}
    \icmlauthor{Liam H Fowl}{equal,last,pli}
    \icmlauthor{Sanjeev Arora}{last,pli}
  \end{icmlauthorlist}

  \icmlaffiliation{pli}{Princeton Language and Intelligence, Princeton University}
  \icmlaffiliation{amazon}{Amazon. This work is independent of and outside of the work at Amazon}

  \icmlcorrespondingauthor{Jui-Hui Chung}{jc1220@princeton.edu}
  \icmlcorrespondingauthor{Ziyang Cai}{zc5794@princeton.edu}
  \icmlcorrespondingauthor{Liam H Fowl}{lf2728@princeton.edu}

  \icmlkeywords{Machine Learning, ICML}

  \vskip 0.3in
]



\printAffiliationsAndNotice{\icmlEqualContribution\textsuperscript{\dag}Joint last authors.}

\begin{abstract}
We introduce \textsc{Goedel-Architect}, an agentic framework for formal
theorem proving in Lean~4 centered on blueprint generation and
refinement. A blueprint is a dependency graph of definitions and
lemmas that builds up to the main theorem. First, \textsc{Goedel-Architect} generates a blueprint of formally stated definitions and lemmas, along with declared dependencies. This blueprint is optionally guided by a natural language proof. Then, a tool-equipped Lean prover component closes each open lemma node in parallel using relevant dependencies. Failed lemmas in turn drive refinement of the global blueprint. This strategy contrasts with other mainstream approaches which use recursive lemma decomposition, and can inefficiently loop on dead-end strategies. Using the open-weight DeepSeek-V4-Flash
(284B-A13B) as the backbone, \textsc{Goedel-Architect} attains
$99.2\%$ pass@$1$ on MiniF2F-test and $75.6\%$ pass@$1$
on PutnamBench. With an optional natural-language
proof seeding the initial blueprint on the harder problems, we
additionally close the remaining two MiniF2F-test problems (reaching
$100\%$), lift PutnamBench to $88.8\%$ ($597/672$), and solve $4/6$
on IMO 2025, $11/12$ on Putnam 2025, and $3/6$ on USAMO 2026. This represents state-of-the-art performance for an open-source pipeline at a price point up to $500\times$ less than comparable open-source pipelines.
\end{abstract}

\section{Introduction}

The last two years have seen rapid progress in the mathematical abilities of frontier AI systems. This progress has been highlighted by several achievements by AI systems such as IMO gold-winning submissions from several frontier labs \cite{deepmind2025imo, wei2025openaiimo}, to solving open Erd\H{o}s problems that had previously deterred even elite math researchers \cite{alexeev2026primitive}, among others. As mathematical capabilities of AI systems continue to develop, the human cost of verification of AI-generated proofs also proportionally grows.

\begin{figure*}[t]
\centering
\tikzset{
    bpU/.style={circle, draw=blue!70!black, fill=blue!22,
                minimum size=4.2mm, inner sep=0pt},
    bpS/.style={circle, draw=green!55!black, fill=green!40,
                minimum size=4.2mm, inner sep=0pt},
    bpN/.style={circle, draw=red!70!black, fill=red!30,
                minimum size=4.2mm, inner sep=0pt},
    bpNew/.style={circle, draw=blue!70!black, fill=blue!22, dashed,
                  line width=0.5pt, minimum size=4.2mm, inner sep=0pt},
    bptU/.style={bpU, font=\tiny},
    bptS/.style={bpS, font=\tiny},
    bparrow/.style={-{Latex[length=1.1mm]}, thin, gray!55!black},
}
\begin{tikzpicture}[
    >={Latex[length=2.4mm]},
    font=\footnotesize,
    every node/.style={align=center},
    stage/.style={rectangle, draw, thick, rounded corners, fill=blue!6,
                  minimum height=1.0cm, minimum width=2.1cm, inner sep=3pt},
    input/.style={rectangle, draw, rounded corners, fill=gray!10,
                  minimum height=0.7cm, minimum width=1.7cm, inner sep=3pt},
    optional/.style={rectangle, draw, dashed, rounded corners, fill=blue!6,
                     minimum height=0.7cm, minimum width=1.9cm, inner sep=3pt},
    bp/.style={rectangle, draw, rounded corners, fill=yellow!15,
               minimum height=1.4cm, minimum width=2.3cm,
               inner sep=3pt, font=\scriptsize},
    diag/.style={rectangle, draw, rounded corners, fill=red!8,
                 minimum height=0.8cm, minimum width=2.3cm,
                 inner sep=2pt, font=\scriptsize},
    success/.style={rectangle, draw, thick, rounded corners, fill=green!14,
                    minimum height=0.9cm, minimum width=1.8cm, inner sep=3pt},
    tool/.style={rectangle, draw, rounded corners, fill=orange!16,
                 minimum height=0.7cm, minimum width=2.0cm, inner sep=3pt,
                 font=\scriptsize},
    edgelabel/.style={font=\scriptsize, fill=white, inner sep=1pt},
]
\node[input, dashed] (informal) at (0,    0.95) {Informal\\statement};
\node[input]    (formal)   at (0,    0)    {Formal\\statement};
\node[optional] (nl)       at (2.7,  0.95) {Natural-language\\proof};
\node[stage]    (bg)       at (5.5,  0)    {Blueprint\\generation};
\node[bp]       (gk)       at (8.5,  0)    {%
\begin{tikzpicture}
\node[bpU]  (a) at (0,    0.30) {};
\node[bpU]  (b) at (0.62, 0)    {};
\node[bpU]  (c) at (0,   -0.30) {};
\node[bptU] (t) at (1.32, 0)    {$T$};
\draw[bparrow] (a)--(b); \draw[bparrow] (c)--(b); \draw[bparrow] (b)--(t);
\end{tikzpicture}\\[-1pt]
Initial blueprint $G_0$};
\node[stage]    (tp)       at (11.5, 0)    {Theorem\\proving};
\node[tool]     (lean)     at (11.5, 1.55) {Lean compiler\\+ Mathlib};
\node[bp]       (gstar)    at (14.4, 0)    {%
\begin{tikzpicture}
\node[bpS]  (a) at (0,    0.30) {};
\node[bpS]  (b) at (0.62, 0)    {};
\node[bpS]  (c) at (0,   -0.30) {};
\node[bptS] (t) at (1.32, 0)    {$T$};
\draw[bparrow] (a)--(b); \draw[bparrow] (c)--(b); \draw[bparrow] (b)--(t);
\end{tikzpicture}\\[-1pt]
Solved blueprint};

\node[success]  (out)      at (14.4, -2.85) {Lean proof\\of $T$};
\node[bp]       (verdict)  at (11.5, -2.85) {%
\begin{tikzpicture}
\node[bpS]  (a) at (0,    0.30) {};
\node[bpU]  (b) at (0.62, 0)    {};
\node[bpN]  (c) at (0,   -0.30) {};
\node[bptS] (t) at (1.32, 0)    {$T$};
\draw[bparrow] (a)--(b); \draw[bparrow] (c)--(b); \draw[bparrow] (b)--(t);
\end{tikzpicture}\\[-1pt]
Result blueprint};
\node[diag]     (wrong)    at (8.5, -2.45) {\textsc{statement\_wrong}\\alter formalization};
\node[diag]     (hard)     at (8.5, -3.35) {\textsc{proof\_too\_hard}\\decompose lemma};
\node[stage]    (br)       at (5.5, -2.85) {Blueprint\\refinement};
\node[bp]       (gnext)    at (1.4, -2.85) {%
\begin{tikzpicture}
\node[bpS]   (a) at (0,    0.30) {};
\node[bpU]   (b) at (0.62, 0.30) {};
\node[bpNew] (c) at (0,   -0.30) {};
\node[bpNew] (d) at (0.62, -0.30) {};
\node[bptS]  (t) at (1.34, 0)    {$T$};
\draw[bparrow] (a)--(b); \draw[bparrow] (c)--(b);
\draw[bparrow] (d)--(b); \draw[bparrow] (b)--(t);
\end{tikzpicture}\\[-1pt]
Revised blueprint $G_{k+1}$};

\draw[->, dashed] (informal) -- (nl);
\draw[->, dashed] (nl.east) -| (bg.north);
\draw[->] (formal) -- (bg);
\draw[->] (bg) -- (gk);
\draw[->] (gk) -- (tp);
\draw[->] (tp) -- (gstar);
\draw[->] (gstar) -- (out);
\draw[<->] (tp.north) -- (lean.south);
\draw[->] (tp) -- (verdict);
\draw[->] (verdict.west) -- (wrong.east);
\draw[->] (verdict.west) -- (hard.east);
\draw[->] (wrong.west) -- (br.east);
\draw[->] (hard.west) -- (br.east);
\draw[->] (br) -- (gnext);
\draw[->] (gnext.north) |- (10.8,-1.4) -- ([xshift=-7mm]tp.south);
\end{tikzpicture}
\caption{Overview of the \textsc{Goedel-Architect} pipeline. \textbf{Inputs.} The formal statement always seeds blueprint generation; an informal statement and a natural-language proof (dashed) form an optional structural guide. \textbf{Blueprint generation.} It emits an initial dependency graph $G_0$ of definitions and lemmas building up to the main theorem $T$ (arrows are declared dependencies); every node starts \emph{unsolved} (blue). Theorem proving dispatches each lemma to a Lean prover in parallel, restricted to its declared parents, with every candidate proof checked against the Lean compiler and Mathlib, and emits a result blueprint marking each node \emph{solved} (green), \emph{unsolved} (blue), or \emph{formally negated} (red). When every node is solved, the solved blueprint assembles into a Lean proof of $T$. \textbf{Refinement loop.} Otherwise, each failing node carries one of two diagnoses: \textsc{statement\_wrong} --- the statement is false under its hypotheses, so the formalization is altered --- raised either by a formally negated (red) node, where the prover verified a counterexample, or by an unsolved (blue) node the prover argues is false but could not disprove; or \textsc{proof\_too\_hard} --- an unsolved (blue) node the prover believes provable but could not chain its parents to, so the lemma is decomposed into helper lemmas. Blueprint refinement consumes these signals and emits the revised blueprint $G_{k+1}$, preserving already-solved nodes as green (their proofs reused while signatures and parents are unchanged, though an upstream edit can revert a dependent to blue) and adding new helper lemmas (dashed). The revised blueprint is re-proven, and the loop continues until every node is solved or the iteration budget is exhausted.}
\label{fig:pipeline}
\end{figure*}
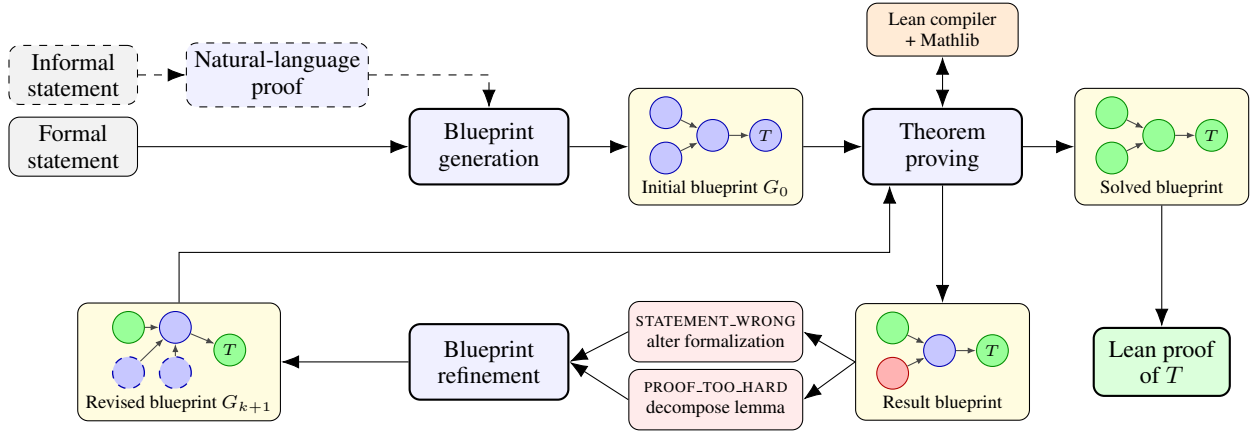

Formal theorem proving, in verifiable languages like Lean \cite{moura2021lean}, offers an appealing proposition for researchers wanting to harness the math abilities of AI systems while also requiring rigor and verifiability of the AI's output. Significant progress has also been made in the vein of formal theorem proving in recent years. Google's AlphaProof \cite{alphaproof} attained silver-winning performance on the 2024 IMO exam, and difficult benchmarks such as PutnamBench \cite{tsoukalas2024putnambench} are now largely tractable for the strongest formal theorem proving systems. Such RL-trained frontier search systems define the upper end of formal-proof capability, albeit with closed weights and reported runs several orders of magnitude above our compute budget.

However, for researchers and enthusiasts, the selection of available, high-quality resources is highly limited. Custom-trained Lean provers are available to academics \cite{lin2025goedel, lin2025goedelv2, ren2025deepseek}, but usually score quite poorly ($<15\%$) on more difficult benchmarks like PutnamBench. Other pipelines and systems can score above $50\%$ on PutnamBench, and some even exceed a $99\%$ solve rate. However, at the time of submission, every system exceeding $50\%$ on PutnamBench either uses \textit{unreleased model or pipeline} \cite{aleph_prover_2025, chen2025seedproverdeepbroadreasoning, chen2025seed}, or requires \textit{expensive frontier model usage} - requiring many thousands of dollars in API credits to solve the slate of questions in PutnamBench \cite{varambally2025hilbert, requena2026minimal}.

To organize this landscape, we group prior work by the role the LLM plays at inference time:
\emph{non-agentic provers} that emit a complete Lean proof in one
shot, \emph{agentic provers} where a single LLM interleaves its
reasoning with calls to the Lean compiler and to retrieval, and
\emph{pipelines} that orchestrate multiple LLM components around
the prover. Where a comparator's backbone or pipeline is
unreleased we mark it \emph{closed}; otherwise the backbone weights
and pipeline code are publicly available (\emph{open}).

\paragraph{Non-agentic provers.}
Open-weight LLMs are fine-tuned on Lean data and used in a single
forward pass: a problem statement goes in, a complete proof comes
out, with no tool use or iteration. The Goedel-Prover series
\cite{lin2025goedel,lin2025goedelv2}, the DeepSeek-Prover line
\cite{xin2024deepseek,xin2024deepseekv15,ren2025deepseek},
Kimina-Prover \cite{wang2025kimina},
Hunyuan-Prover \cite{li2024hunyuanprover}, and
TheoremLLaMA \cite{wang2024theoremllama} all follow this template
and release their weights (open). Lean-STaR \cite{lin2024lean}
additionally interleaves chain-of-thought tokens between proof
steps but still emits the proof in one shot. These provers are
the closest open comparators on MiniF2F; their pass-rates on
Putnam-class problems sit in the single digits, motivating the
agentic and pipeline approaches below.

\paragraph{Agentic provers.}
A single LLM interleaves its reasoning with calls to the Lean
compiler (typecheck errors, goal states) and to a Mathlib
retrieval service, adjusting subsequent steps in response.
AxProverBase \cite{requena2026minimal} and Numina-Lean-Agent
\cite{liu2026numina} pair this template with a proprietary
frontier backbone --- Claude Opus 4.5 in both cases (closed
backbone, open pipeline) --- achieving strong results on
PutnamBench and Putnam 2025 than non-agentic provers.


\paragraph{Pipelines.}
Another approach wraps the prover (agentic or not) in an
orchestrated multi-stage system with explicit decomposition,
refinement, or sketching. Hilbert \cite{varambally2025hilbert}
recursively subdivides goals and proves leaves, on a closed
Gemini 2.5 Pro backbone (open pipeline); Seed-Prover and
Seed-Prover 1.5
\cite{chen2025seedproverdeepbroadreasoning,chen2025seed} combine
sketching with iterative refinement on a closed backbone and
closed pipeline; LongCat-Flash-Prover \cite{wang2026longcat}
releases an open backbone but a closed pipeline; the Aleph
prover \cite{aleph_prover_2025} is closed on both axes. Earlier,
Draft-Sketch-Prove \cite{jiang2022draft} established the template
of letting an LLM draft an informal proof, translating it to a
formal proof sketch with placeholder goals, and using a
downstream prover to fill them in --- the original pipeline that
seeds formal proving with an informal scaffold.

In this work, we introduce \textsc{Goedel-Architect}, which
establishes a new Pareto frontier for formal theorem proving by
delivering state-of-the-art performance for its compute class,
rivaling massive proprietary systems while utilizing a highly
efficient, open-weight backbone. \textsc{Goedel-Architect} sits
in the pipeline category and differs on two dimensions. First,
its central mechanism is a global dependency-graph \emph{blueprint}
rewritten between iterations, rather than a recursion tree built
top-down; this lets parallel proof attempts share context and
lets refinement act on the whole strategy at once. Second, both
backbone and pipeline are open, at a per-problem cost of
$\sim\$0.44$ versus $\sim\$244$ for the next-best open pipeline
(Table~\ref{tab:putnam-spend}). The optional natural-language
proof seed (Section~\ref{sec:nl-helps}) adapts the
Draft-Sketch-Prove idea to this setting: the informal proof lands
as a graph of named sub-lemmas the rest of the pipeline can
refine, rather than as a flat sketch the prover either fills or
discards.

\section{Goedel-Architect}
\label{sec:pipeline}


In this section, we describe the \textsc{Goedel-Architect} pipeline. The core innovation of our pipeline is organized around a
\emph{blueprint}: a dependency graph of definitions and lemmas that
build up to the target theorem. At a high level, the pipeline begins with an initial blueprint generation, optionally guided by a natural-language proof. The pipeline then iterates between Lean theorem proving of the blueprint nodes, and global blueprint refinement. Figure \ref{fig:pipeline} offers a detailed schematic of our pipeline.

\subsection{Blueprint generation}
\label{sec:blueprint-generation}

The blueprint generation stage receives the formal statement of the
target theorem and emits a dependency graph as a single Lean file.
Each node is a formally stated definition or lemma. Each lemma node also
declares which other nodes its proof is allowed to rely on; these
declared dependencies become the edges of the graph, recorded and
validated by the LeanArchitect Lean package
\citep{leanarchitect}. The target theorem is the unique sink of the
graph and keeps the signature of the original formal statement. Lemma bodies are left unproved at this
stage. The model iterates against the Lean compiler so that the
emitted file parses, every node is well-typed, and the graph is
well-formed (acyclic, with every node reachable from the target).


\paragraph{Natural-language proof guidance.}
By default, no natural-language proof is provided during the blueprint generation phase. However, for more difficult problems,
blueprint generation can be optionally guided by a natural-language proof ---
either produced by a stronger model or a sophisticated
natural-language proving pipeline, or supplied as an official
solution (e.g., from a problem set or competition write-up). By
design, this natural-language proof contains only the informal
mathematical argument and is not Lean-aware, so that any strong
informal prover can be used to generate it. The blueprint generator
consumes that proof as a structural guide for the dependency graph.
We find that the resulting blueprint reflects the strategy in the
natural-language proof closely: some strategies are easier to
realize formally than others that are easier to state informally.

\subsection{Theorem proving}
\label{sec:theorem-proving}

Each lemma in the blueprint is dispatched to a Lean theorem prover.
The prover sees only the lemma it is proving and the definitions and
lemmas it declared as dependencies, not the rest of the graph; the
declared dependencies are presented as available facts whose
signatures the prover may invoke by name. Lemmas are proved in
parallel. The prover has access to the Lean compiler and a Mathlib
retrieval tool, and may call them iteratively until it either closes
the goal or exhausts its per-lemma budget. When the prover gives up
on a lemma, it returns a structured diagnosis recording what it
attempted and where it believes the gap lies; when it produces a
compiler-corroborated counterexample, it can register a proof of the
negated statement in lieu of the original. These per-lemma signals
are the input to blueprint refinement.


\begin{table*}[t]
\centering
\caption{Benchmark comparison across MiniF2F-test, PutnamBench, IMO
2025, Putnam 2025, and USAMO 2026. We report \textsc{Goedel-Architect}
in two modes: the default pass@$1$ pipeline, and
\textsc{Goedel-Architect} (+ NL), which augments the pipeline with a
natural-language proof sketch used to seed additional blueprint
attempts on problems left open by the default mode. Results for
other provers are taken from their
respective papers or the dataset papers: Goedel-Prover-V2~\cite{lin2025goedelv2},
LongCat-Flash-Prover~\cite{wang2026longcat},
Seed-Prover~\cite{chen2025seedproverdeepbroadreasoning},
Seed-Prover 1.5~\cite{chen2025seed},
Hilbert~\cite{varambally2025hilbert},
Numina-Lean-Agent~\cite{liu2026numina},
AxProverBase~\cite{requena2026minimal}.}
\label{tab:benchmarks}
\vskip 0.1in
\renewcommand{\arraystretch}{1.3}
\begin{small}
\centering
\begin{tabular}{l r@{}c@{}l r@{}c@{}l r@{}c@{}l r@{}c@{}l r@{}c@{}l}
\toprule
Model & \multicolumn{3}{c}{MiniF2F-test} & \multicolumn{3}{c}{PutnamBench} & \multicolumn{3}{c}{IMO 2025} & \multicolumn{3}{c}{Putnam 2025} & \multicolumn{3}{c}{USAMO 2026} \\
\midrule
Goedel-Prover-V2          & $92.6\%$ & {\,@\,} & $1024$ & $13.0\%$ & {\,@\,} & $184$ & \multicolumn{3}{c}{--} & \multicolumn{3}{c}{--} & \multicolumn{3}{c}{--} \\
LongCat-Flash-Prover      & $97.1\%$ & {\,@\,} & $72$ & $41.5\%$ & {\,@\,} & $118$ & \multicolumn{3}{c}{--} & \multicolumn{3}{c}{--} & \multicolumn{3}{c}{--} \\
Seed-Prover               & $99.6\%$ & & & $50.4\%$ & & & \multicolumn{3}{c}{$5/6$} & \multicolumn{3}{c}{--} & \multicolumn{3}{c}{--} \\
Seed-Prover 1.5           & \multicolumn{3}{c}{--} & $87.9\%$ & & & \multicolumn{3}{c}{$5/6$} & \multicolumn{3}{c}{$11/12$} & \multicolumn{3}{c}{--} \\
Hilbert                   & $99.2\%$ & & & $70.0\%$ & {\,@\,} & $1840$ & \multicolumn{3}{c}{--} & \multicolumn{3}{c}{--} & \multicolumn{3}{c}{--} \\
Numina-Lean-Agent         & \multicolumn{3}{c}{--} & \multicolumn{3}{c}{--} & \multicolumn{3}{c}{--} & \multicolumn{3}{c}{$12/12$} & \multicolumn{3}{c}{--} \\
AxProverBase              & \multicolumn{3}{c}{--} & $54.7\%$ & {\,@\,} & $1$ & \multicolumn{3}{c}{--} & \multicolumn{3}{c}{--} & \multicolumn{3}{c}{--} \\
\midrule
\textbf{Goedel-Architect} & $99.2\%$ & {\,@\,} & $1$ & $75.6\%$ & {\,@\,} & $1$ & \multicolumn{3}{c}{--} & \multicolumn{3}{c}{--} & \multicolumn{3}{c}{--} \\
\textbf{Goedel-Architect} (+ NL) & $100\%$ & & & $88.8\%$ & {\,@\,} & $4$ & \multicolumn{3}{c}{$4/6$} & \multicolumn{3}{c}{$11/12$} & \multicolumn{3}{c}{$3/6$} \\
\bottomrule
\end{tabular}
\end{small}
\end{table*}

\subsection{Blueprint refinement}
\label{sec:blueprint-refinement}

If any lemma is unproved after a proving pass, the blueprint
refinement stage rewrites the graph around the failures. A refinement
model reads the per-lemma traces, marks each lemma as proved or
unproved (carrying the prover's diagnosis and any formal disproof for
unproved lemmas), and emits a revised graph. Typical refinements
include decomposing a hard lemma into intermediate helper lemmas,
rewiring dependencies so a lemma has access to results it needs, and
repairing or dropping the statement of a lemma the prover argued was
false. Lemmas that the prior pass proved are preserved with their
signatures intact, so the proving budget already spent on them is not
discarded. The refined graph is handed to another proving pass, and
the loop continues until every lemma is proved or the iteration
budget is reached.


\section{Experiments}
\label{sec:results}

\subsection{Benchmarks}
\label{sec:results-benchmarks-desc}

We test \textsc{Goedel-Architect} on five Lean benchmarks that comprise high-school and undergraduate competition mathematics and contamination-free olympiad problems, among others. \textbf{MiniF2F-test}
\cite{zheng2021minif2f} is the standard $244$-problem suite of
high-school competition problems --- algebra, number theory, and
inequalities drawn from AMC, AIME, and the IMO --- and is the easiest
of the five, now nearly saturated by the strongest provers.
\textbf{PutnamBench}~\cite{tsoukalas2024putnambench} raises the
difficulty to undergraduate-level competition mathematics, with $672$
Lean formalizations of past William Lowell Putnam Competition problems
spanning analysis, algebra, combinatorics, and number theory; it is
a benchmark on which current provers still differ significantly.

Additionally, we report results on three small, recent competition sets that probe
performance on fresh problems. \textbf{IMO 2025} and \textbf{USAMO
2026} are six-problem pre-university olympiad exams at the hardest
competition tier, and \textbf{Putnam 2025} is the twelve-problem
(A1--A6, B1--B6) undergraduate exam from the aforementioned Putnam competition. Their formalized statements come
from different sources: for IMO 2025 we use the formalizations released
with the Seed-Prover paper~\cite{chen2025seed}; for Putnam 2025 we use
those in PutnamBench~\cite{tsoukalas2024putnambench}; and for USAMO
2026, which postdates the training cutoff of every model in our
pipeline and so serves as a contamination-free benchmark, we formalize
the statements ourselves with the help of Claude Opus~4.7 \cite{anthropic2026claudeopus47}.

\subsection{Main results}
\label{sec:results-benchmarks}

Table~\ref{tab:benchmarks} reports \textsc{Goedel-Architect} against
the strongest publicly reported provers across the five benchmarks:
MiniF2F-test, PutnamBench, IMO 2025, Putnam 2025, and USAMO 2026. Not every pipeline tested on every benchmark, and accordingly, corresponding results are left blank.

\paragraph{MiniF2F.} On MiniF2F-test~\cite{zheng2021minif2f},
\textsc{Goedel-Architect} solves $242/244$ problems at pass@$1$
($99.2\%$) without using any natural-language proof as guidance.
Pass@$1$ here is measured at the
\emph{pipeline} level rather than at the prover level: each problem
gets exactly one blueprint generation, and that single blueprint is
then refined up to $8$ times.

The remaining two problems --- IMO 1984 P6 and IMO Shortlist 2007
Algebra P6 --- are not closed at pass@$1$. We close them in a
separate, more expensive regime: several blueprint attempts seeded
with a natural-language proof generated by Gemini~3.1~Pro \cite{google2026gemini31pro}, with the
same proving and refinement pipeline. Notably,
Seed-Prover~\cite{chen2025seedproverdeepbroadreasoning} was able to
solve all problems in MiniF2F-test except for IMO Shortlist 2007
Algebra P6. To our knowledge, \textsc{Goedel-Architect} is the first
Lean prover to close all $244$ problems on MiniF2F-test, albeit with
natural-language guidance and additional samples on two of them; the
previous best on the test split was Seed-Prover at $243/244$.


\begin{table}[h]
\centering
\caption{PutnamBench cost comparison between Goedel-Architect (DeepSeek-V4-Flash), Hilbert (Gemini 2.5 Pro), and AxProverBase. Total cost and average cost per question (denoted Avg. cost / Q) are reported. Note that the Hilbert's cost is taken from pass@1 numbers, and is an \textit{underestimate} of their actual spend.}
\label{tab:putnam-spend}
\vskip 0.1in
\begin{small}
\begin{tabular*}{\columnwidth}{l@{\extracolsep{\fill}}rrr}
\toprule
Metric & Goedel-Arch. & Hilbert* & AxProver \\
\midrule
Total spend                       & \$294  & $\sim$\$163k & \$8{,}467 \\
Avg.\ cost / Q (all)              & \$0.44 & $\sim$\$244  & \$12.60   \\
Avg.\ cost / \emph{solved} Q      & \$0.21 & ---          & ---       \\
Avg.\ cost / \emph{unsolved} Q    & \$1.14 & ---          & ---       \\
\bottomrule
\end{tabular*}
\end{small}
\end{table}

\begin{figure}[t]
\centering
\includegraphics[width=\linewidth]{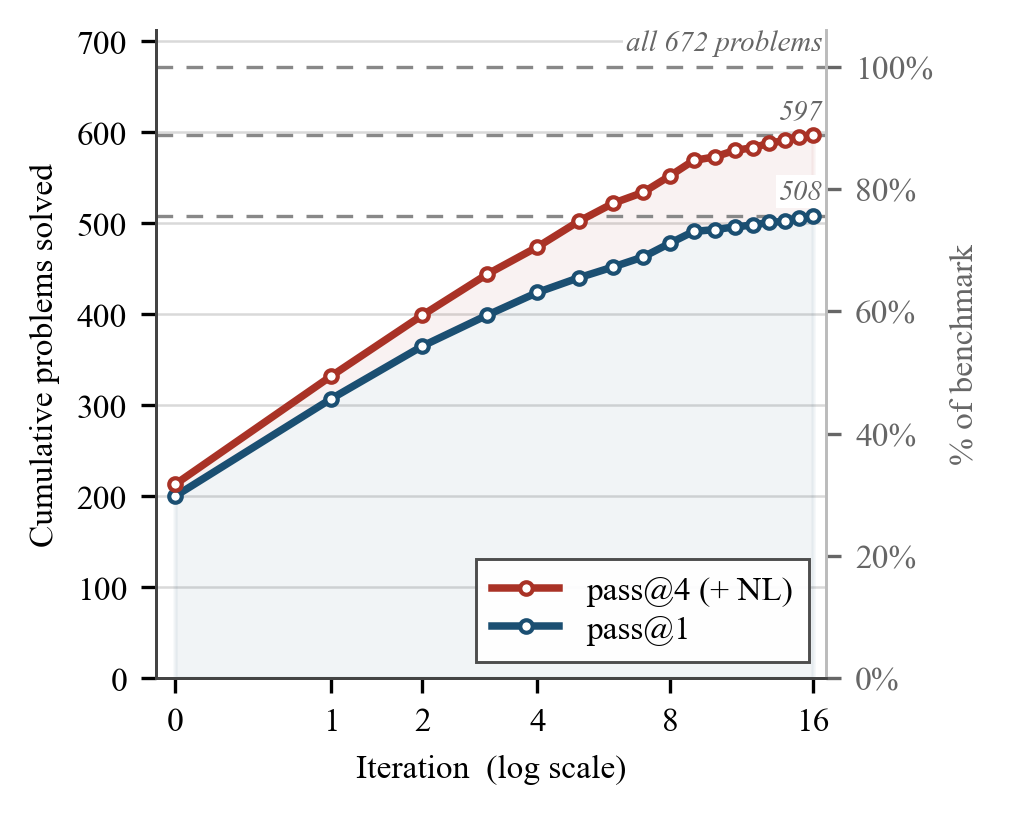}
\caption{Compute scaling on PutnamBench. Cumulative problems solved
by \textsc{Goedel-Architect} (left axis; right axis as \% of the
$672$-problem benchmark) against the number of blueprint-refinement
iterations on a log scale, using the open-weight DeepSeek-V4-Flash
backbone. The pass@$1$ curve uses only the default pipeline; the
pass@$4$ (+ NL) curve is the pass@$4$ NL effort. The initial blueprint (iteration~$0$)
alone closes $200$ problems; each subsequent refinement pass adds
more, reaching $508$ ($75.6\%$) at pass@$1$ and $597$ ($88.8\%$) at
pass@$4$ (+ NL) by iteration~$16$. Solve count grows roughly
log-linearly with refinement compute.}
\label{fig:solved-vs-iter}
\end{figure}

\paragraph{PutnamBench.} Performance differences become more visible on the more difficult PutnamBench. At
pass@$1$, \textsc{Goedel-Architect} solves $75.6\%$ of problems,
exceeding Hilbert's $70.0\%$ at pass@$1840$ and AxProverBase's
$54.7\%$ at pass@$1$ on the same single-sample budget. As in the
MiniF2F result above, pass@$1$ is measured at the pipeline level ---
one blueprint generation per problem --- but here the single
blueprint is refined up to $16$ times. Potentially just as interesting as the score differences are the \textit{backbone} differences across pipelines: AxProverBase runs
on proprietary Claude Opus~4.5 \cite{anthropic2025claudeopus45}, and Hilbert on proprietary Gemini~2.5~Pro \cite{comanici2025gemini}, whereas \textsc{Goedel-Architect} uses the open-weight
DeepSeek-V4-Flash (284B-A13B), whose inference cost is a small
fraction of either frontier API. The score also scales predictably
with refinement budget: as illustrated in
Figure~\ref{fig:solved-vs-iter}, investing more compute in blueprint
refinement iterations yields a roughly log-linear increase in
\textsc{Goedel-Architect}'s solve rate, rising from $200$ problems
($29.8\%$) at the initial blueprint to $508$ ($75.6\%$) by iteration
$16$. Note we confirm that backbone differences are not responsible for our improved performance and efficiency in Section \ref{sec:results-comparison}.

Additionally, we find that the combination of backbone model, and efficient pipeline design reduces cost by close to two orders of magnitude. In Table \ref{tab:putnam-spend} we compare our spend on PutnamBench to that of the next best open-source pipeline (Hilbert). We complete our evaluation of the $672$ PutnamBench questions with just $\$294$ spent on API calls, whereas Hilbert uses close to $\$170{,}000$ in Gemini credits just to finish a single run of the benchmark! And this is likely a significant underestimate of total spend as their compute numbers are presented only for successful proofs, and only for pass@1, whereas their full PutnamBench numbers utilize pass@1840.

We can push performance on PutnamBench further by augmenting the
blueprint with natural-language guidance. The natural-language proof
sketches that seed the blueprint are produced by a separate
generation/correction pipeline running on DeepSeek-V4-Flash or
DeepSeek-V4-Pro, keeping the entire stack open-source. At pass@$4$
with this NL guidance, \textsc{Goedel-Architect} closes $597/672$
problems ($88.8\%$), an absolute $+13.2\%$ over the default
pipeline. The entire pass@$4$ effort cost $\$985.67$ in API calls
(including all unsuccessful attempts), or $\sim\$1.65$ per problem
solved.

\paragraph{IMO 2025.} We take the formalized problem statements for
IMO 2025 from the Seed-Prover paper~\cite{chen2025seed}.
\textsc{Goedel-Architect} solves $4/6$ problems (P1, P3, P4, P5),
compared to Seed-Prover 1.5's $5/6$. The extra problem in
Seed-Prover's tally is P2, a geometry problem that is not accessible
to a general-purpose Lean prover; Seed-Prover handles it with a
dedicated engine. \textsc{Goedel-Architect} matches Seed-Prover 1.5
on the remaining problems, and is the only open-weight system to
reach this tier on IMO 2025 in our comparison.

Of the four solved problems, P5 is closed without any
natural-language proof as input: blueprint generation runs directly
from the formal statement. For P1, P3, and P4 we use Gemini~3.1~Pro
to supply an initial natural-language proof, which seeds the first
blueprint generation pass as a structural guide; the rest of the
pipeline (proving and refinement) is unchanged.

\paragraph{Putnam 2025.} The formalized problem statements for Putnam
2025 are taken from PutnamBench~\cite{tsoukalas2024putnambench}.
\textsc{Goedel-Architect} solves $11/12$ problems, matching
Seed-Prover 1.5 and again standing alone among open-weight provers.
Numina-Lean-Agent~\cite{liu2026numina} clears $12/12$ using the
proprietary Claude Opus~4.5.

Of the eleven solved problems, A1, A2, A4, A6, B2, B3, and B5 are
closed without any natural-language proof as input: blueprint
generation runs directly from the formal statement, and these are
Gsolved as part of the same pass@$1$ run we report on PutnamBench. For
A3, B1, B4, and B6 we use Gemini~3.1~Pro to supply an initial
natural-language proof, which seeds the first blueprint generation
pass as a structural guide; the rest of the pipeline (proving and
refinement) is unchanged.

\paragraph{USAMO 2026.} USAMO 2026 is recent enough to postdate the
training cutoff of every model in our pipeline, so it serves as a
contamination-free benchmark: neither the problem statements nor
their solutions could have been memorized during training. We
formalize the USAMO 2026 problem statements ourselves, with the help
of Claude Opus~4.7. \textsc{Goedel-Architect} solves $3/6$
problems (P1, P4, P6). For all
three we use Gemini~3.1~Pro to supply an initial natural-language
proof, which seeds the first blueprint generation pass as a
structural guide; the rest of the pipeline (proving and refinement)
is unchanged.

\section{Other Pipeline Features}
\label{sec:pipeline_features}
In this section, we describe some of auxiliary pipeline features, and their impact on the entire pipeline, as well as studying some cases that highlight their utility. 

\subsection{Effectiveness of base model and tool use}
\label{sec:results-comparison}

To characterize what the base model can do on its own, we evaluate
DeepSeek-V4-Flash on MiniF2F-test (244 problems)
\cite{zheng2021minif2f} as a single agent under two regimes. In
\emph{direct inference}, the model is given the problem statement and
emits a complete Lean proof in one shot. In
\emph{tool-integrated reasoning}, the model interleaves reasoning with
calls to the Lean verifier during generation and to a semantic Mathlib
search service for lemma retrieval, observing typecheck errors and
goal states and adjusting subsequent steps accordingly. We
additionally run Hilbert~\cite{varambally2025hilbert}, a recursive
informal-reasoning-with-formal-verification pipeline, with both LLM
roles in Hilbert (the informal reasoner and the leaf prover) swapped to
our same backbone. We reproduce Hilbert's published algorithm verbatim ---
recursive subgoal decomposition, sketch generation and verification,
per leaf error correction, and depth bounded recursion --- with two
thin adapters that route the verifier to our Lean gateway and Mathlib
retrieval to our semantic search service. We also adapt Hilbert's pipeline to use our node prover to further distinguish pipeline features.

In Table \ref{tab:base-model}, we see that our pipeline design provides an even bigger boost in performance compared to Hilbert when we control for backbone model. We also observe that on easier problems, tool-integrated reasoning (TIR) naturally yields relatively high performance on minif2f. However, Figure \ref{fig:solved-vs-iter} shows that TIR is alone insufficient to achieve our performance numbers on the more difficult PutnamBench.

\begin{table}[h]
\centering
\caption{Base-model effectiveness on the same DeepSeek-V4-Flash backbone.
Direct inference and tool-integrated reasoning (TIR) isolate the two single-agent
regimes; the Hilbert rows port Hilbert's recursive-decomposition
algorithm onto the same backbone, with and without the TIR prover swapped in.
PutnamBench numbers are over a random 200-problem subset due to control compute.}
\label{tab:base-model}
\vskip 0.1in
\begin{small}
\begin{tabular}{lr@{ @ }lr@{ @ }l}
\toprule
Method & \multicolumn{2}{c}{MiniF2F-test} & \multicolumn{2}{c}{PutnamBench} \\
\midrule
Direct inference            & $67.6\%$ & $32$ & $6.5\%$  & $32$ \\
Tool-integrated reasoning   & $97.1\%$ & $32$ & $54.5\%$ & $32$ \\
Hilbert                     & $83.6\%$ & $1$  & \multicolumn{2}{c}{--} \\
Hilbert (w/ TIR)            & $84.4\%$ & $1$  & \multicolumn{2}{c}{--} \\
\bottomrule
\end{tabular}
\end{small}
\end{table}

\begin{figure}[t]
\centering
\includegraphics[width=\linewidth]{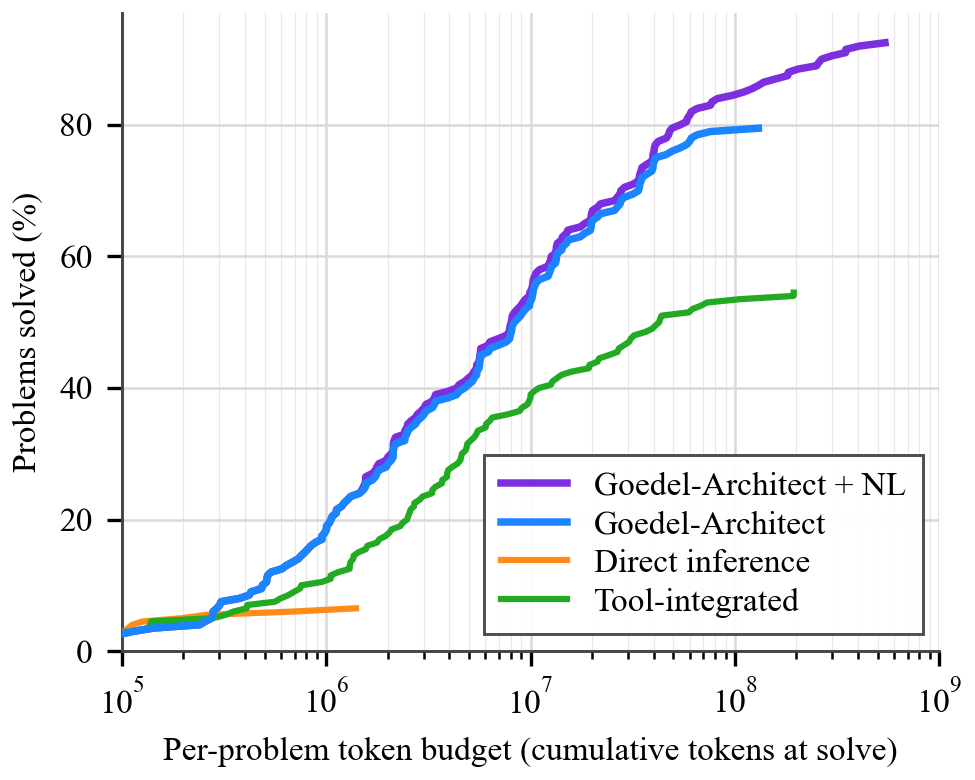}
\caption{Per-problem solve cost on PutnamBench. All three systems
share the same DeepSeek-V4-Flash backbone and are evaluated on the
\emph{same} random subset of $N{=}200$ PutnamBench problems --- those
attempted by all three systems. For each problem we record the
cumulative tokens (input $+$ output) spent on it at the moment it was
first verifiably solved, and plot the fraction of the $200$ problems
solved within a given per-problem token budget. The
tool-integrated agentic baseline reaches $10^8$ tokens per problem in
its tail: with $32$ multi-turn samples per problem under early-stop, a
single hard agentic attempt can cost more than
\textsc{Goedel-Architect}'s entire refinement sweep on the same
problem.}
\label{fig:tokens-vs-passrate}
\end{figure}

Figure~\ref{fig:tokens-vs-passrate} makes the per-problem compute
story concrete. On the common $200$-problem subset,
\textsc{Goedel-Architect} reaches $50\%$ solved at roughly
$7\text{M}$ tokens per problem and climbs to $76.0\%$ ($152/200$) by
its eighth refinement iteration, while the tool-integrated agentic
baseline on the same backbone needs several times as many tokens to
reach a comparable pass rate and tops out at $54.5\%$ ($109/200$).
The no-tools direct-inference baseline never escapes single-digit
pass rates ($6.5\%$, $13/200$).

\subsection{When natural-language guidance helps}
\label{sec:nl-helps}

We study nine problems that our pipeline does not close without
natural-language guidance: the two MiniF2F-test problems our
pass@$1$ effort leaves open (IMO 1984 P6 and IMO Shortlist 2007
Algebra P6), IMO~2025 P1, P3, and P4, and Putnam~2025 A3, B1, B4,
and B6. For each, we re-run the pipeline \emph{with natural
language}, seeding the initial blueprint with a Gemini~3.1~Pro
natural-language proof while holding the backbone, iteration
budget, and per-stage settings identical to the formal pipeline
and issuing a comparable number of runs,
so the comparison is fair. With this guidance, all nine close.

This is not pass@$1$ in the sample-the-prover sense: each
pipeline invocation already runs the iteration loop to the budget
described in Section~\ref{sec:blueprint-refinement}, and we ran
the pipeline 4 to 12 times (median 6) on each of these problems
in the without-natural-language setting without a single success. We cannot rule out that
significantly more compute in the without-natural-language setting
would eventually close some of them --- but at the compute budget
we report, the contrast between 0 successes without natural language
and 1 to 4 successes with natural language over comparable sample
budgets is consistent across all nine problems. We read this as
evidence that the natural-language proof acts as a structural
scaffold for the initial blueprint: on problems with non-local
structure (cyclic sums, parity or divisibility chains,
tile-counting arguments), deriving the lemma dependency graph from
the formal statement alone is the bottleneck, and seeding the
initial blueprint with an explicit human-style proof anchors the
strategy and lets the refinement loop converge. In short, on these
problems natural-language guidance is decisive.

\subsection{Negated sub-lemmas as proof-strategy diagnostics}
\label{sec:negation-case-study}

When the blueprint proposes a sub-lemma that is actually false, our
per-lemma prover closes a proof of its \emph{negation} instead of the
lemma itself (Section~\ref{sec:theorem-proving}). The refinement loop
treats this as a diagnostic signal: alongside the disproof it records a
short reflection --- a diagnosis, the counterexample that kills the
claim, and a suggested fix --- and feeds that reflection into the next
iteration's blueprint-revision step, which edits the offending node and
the lemmas that depend on it. We see this happen routinely on
PutnamBench: across the full sixteen-iteration refinement sweep,
sub-lemma negations trigger on $292$ of the $672$ problems, with
$1$ to $4$ negated nodes per problem. Two examples show how the revision consumes
the signal.

We discuss illustrative examples of this phenomenon in Appendix \ref{app:negated_nodes}. In both cases here, the negation channel converts a dead end into a
localized, machine-checked edit. A false sub-lemma does not merely fail
to prove; it yields a specific diagnosis --- a missing hypothesis here,
a representational confusion there --- that the next revision step
turns into a concrete change to the blueprint, repairing or re-routing
exactly the broken intermediate claim instead of discarding and
resampling the whole proof.

\subsection{Forfeited proofs as decomposition proposals}
\label{sec:forfeit-case-study}

The negation channel above fires only when the prover can verify a
counterexample. A more common outcome is that a node simply runs out of
budget: the prover can neither close the lemma nor disprove it within
its turn and token limit. Rather than report a bare failure, the prover
is then required to write a structured post-mortem --- a \emph{forfeit}
--- in three parts: a diagnosis of either the statement being wrong or
the proof being too hard; a forensic analysis of what it tried and
where the proof stalled; and a suggested fix. A statement-is-wrong
forfeit flags a lemma the prover suspects is false but could not
actually disprove, and recommends weakening or strengthening it. A
proof-is-too-hard forfeit asserts the lemma is true but out of reach in
one shot, and proposes a \emph{lemma decomposition}: a set of
named helper lemmas arranged so that each is easy given its parents and
the original goal becomes routine given the helpers. As with the
disproof reflection, this text is injected into the next iteration's
blueprint-revision step, which uses it to split or repair the node ---
turning a goal that failed as a single monolith into a sub-graph the
prover can actually discharge.


A case study for this functionality is discussed in Appendix \ref{app:subsec_forefeit}. This is the constructive counterpart to the negation channel. A
disproof tells the loop a node is dead and must be removed; a forfeit
tells it the node is sound but mis-sized, and hands the next revision a
ready-made plan for cutting it into provable pieces.

\clearpage
\newpage
\bibliographystyle{icml2026}
\bibliography{main}

\onecolumn
\appendix

\section{Compute details}
\label{app:compute}
\paragraph{blueprint generation}
By default, the blueprint generation phase emits a single blueprint
to work with. The model's maximum total token length is set to
$262{,}144$ tokens, and it retries up to $8$ times if it fails to
produce a compiling Lean~4 blueprint.

\paragraph{Lean proving}
By default, the prover's maximum total token length is set to
$65{,}536$ tokens, and each node retries up to $4$ times.

\paragraph{Blueprint revising}
By default, the refinement model's maximum total token length is set
to $262{,}144$ tokens, each refinement step retries up to $8$ times,
and the total number of refinement iterations within a single pipeline
pass is set to $8$.

\section{Case studies}
\label{app:case_studies}
Here we include a few interesting case studies to illustrate the functionality/operation of different parts of our pipeline. 

\subsection{Negated nodes}
\label{app:negated_nodes}
\paragraph{Missing hypothesis (Putnam 1971 A6).}
The problem asks: if $n^c$ is an integer for every positive integer
$n$, show that $c$ is a nonnegative integer. The blueprint proposed an
auxiliary characterization claiming that every non-decreasing,
completely multiplicative integer-valued function on the naturals must
be a power function $n \mapsto n^k$. The prover disproved it with the
constant zero function, which is multiplicative and trivially
non-decreasing but is not a power function: a power function takes the
value $1$ at $n = 1$, whereas the zero function takes $0$. The recorded
diagnosis isolates the gap precisely --- complete multiplicativity only
forces $f(1) = f(1)^2$, so $f(1)$ may be $0$ or $1$, and monotonicity
does not exclude the zero solution --- and proposes the fix of adding
the hypothesis $f(1) = 1$. The next iteration's revision adopts this
fix directly: it rewrites the lemma with the extra hypothesis
$f(1) = 1$ and re-derives the surrounding dependency graph around the
strengthened statement, rather than resampling the same false claim.

\paragraph{Representational trap (Putnam 1989 A6).}
The problem reasons about binary representations of integers. The
blueprint proposed an auxiliary identity stating that multiplying a
number by two appends a zero to the end of its binary expansion. The
prover disproved it at $n = 5$: the library lists binary digits
least-significant-bit first, so multiplying by two \emph{prepends} a
zero rather than appending one, and the two sides disagree at every
position. The recorded diagnosis names the orientation confusion and
supplies the corrected identity --- a leading zero rather than a
trailing one. The next iteration's revision propagates this through the
proof: it replaces the false identity with the corrected prepend form,
pushes the same correction into the dependent ``multiply by four''
lemma, and drops the chain of downstream nodes that had been built on
the wrong orientation, substituting helper lemmas consistent with the
fixed convention.

\subsection{Forefeited proofs}
\label{app:subsec_forefeit}
\paragraph{Case split on a polynomial (Putnam 1985 B1).}
The crux lemma asks to show that a monic degree-five polynomial whose
roots are five distinct integers cannot have at most two nonzero
coefficients. The prover spent its budget on low-level rewriting and
never reached the case analysis, so it forfeited, diagnosing the proof
as too hard rather than the statement as wrong. Its analysis nonetheless laid out
the right argument in prose: if every lower coefficient vanishes the
polynomial is $X^5$, whose only root is $0$ with multiplicity five,
contradicting distinctness; otherwise exactly one lower coefficient is
nonzero, giving $X^5 + a X^k$, and each exponent fails for its own
reason --- $X^2$ divides the polynomial when $k \ge 2$, forcing a
repeated root; $X^5 + a$ has at most one integer root; and $X^5 + aX$
factors as $X(X^4 + a)$, with at most three. The suggested fix turned
that prose into a list of helper lemmas, one per case. The next
iteration's revision adopted the decomposition essentially verbatim: it
replaced the single failing node with the proposed case-split lemmas
(the all-coefficients-zero case, the $k \ge 2$, $k = 0$, and $k = 1$
cases, and the supporting degree and coefficient facts) and rewrote the
crux lemma to combine them. Every new node closed on that pass, and
with the crux discharged the full problem was solved --- a goal that had
been unprovable as a monolith one iteration earlier.

\section{System prompts}
\label{app:prompts}

This appendix reproduces the system prompts driving the three stages
of the pipeline: blueprint generation, theorem proving, and blueprint
refinement. User prompts (problem-specific inputs) are omitted; only
the cached behavioral prompts are shown.

\subsection{Blueprint generation}
\label{app:prompt-skeleton}

\begin{lstlisting}[style=prompt]
## Task
You are a Lean 4 formalizer producing a dependency graph decomposition for a Lean theorem. The input is the targeted Lean theorem signature. Design a dependency graph of named Definitions, Lemmas, and exactly one Theorem (the main target), then translate the graph into one Lean 4 file in which every node is a `@[blueprint]`-annotated declaration. You do not prove anything in this stage -- every theorem and lemma body is `:= by sorry_using [...]`.

## Decomposition guidelines
Plan a graph that captures the structure of the proof. Use Definitions for any helper functions, sets, structures, or notation the proof needs. Use Lemmas for intermediate facts that require justification. Use the Theorem for the final claim -- its name MUST equal the targeted theorem identifier given in the user prompt.

Each Lemma should be (nearly) trivial once its parent nodes are taken as given: it should require at most 1-2 new logical ideas beyond its declared dependencies and its own inlined premises. If a step needs more, split it into intermediate lemmas -- use as many components as the proof requires. Independent branches stay independent: if two parts of the proof do not share reasoning, their lemmas should not depend on each other.

Every natural language `statement` field is a closed, typed, standalone proposition: every variable carries an explicit quantifier and domain; every hypothesis the proof uses appears as a premise. Do not reach into ambient context -- restate every theorem-level typing and hypothesis your lemma uses. Every natural language `proof` field is a complete sketch citing each declared dep by backticked name (e.g. "by `lemma_a`", "from `def_b`"); show every key equation, and do not write "by algebra", "obviously", or "one can check".

## Mapping graph nodes to Lean declarations
Emit each node of your decomposition directly as a `@[blueprint ...]`-annotated Lean declaration. Use `snake_case` identifiers derived from content (`k_expansion`, `p_at_101`), not position (`lemma_1`); names must be unique within the file.

- For a Definition, emit:
    @[blueprint (statement := /-- natural language description of what's being defined -/)]
    def name (binders) : type := body
  (or `noncomputable def`, `abbrev`, `structure`, `instance` as fits.) Definitions get a real Lean body, not `sorry_using`.
- For a Lemma or Theorem, emit:
    @[blueprint
      (statement := /-- closed, typed, standalone natural language proposition -/)
      (proof := /-- complete natural language sketch citing parent declarations by backticked name -/)]
    lemma|theorem name (binders) : conclusion := by sorry_using [p1, p2, ...]
  where `sorry_using [...]` lists each parent declaration as a bare Lean identifier (or `sorry_using []` if it has no parents).
- The main Theorem's `name` MUST equal the targeted theorem identifier given in the user prompt, and you must emit it with the original Lean signature (same binders, same conclusion). Do not retype the statement informally.
- Declare nodes in topological order: Definitions first, then Lemmas in dependency order, then the main Theorem last.

## Tool use
Use `lean_compile` to verify the skeleton. Before Lean is invoked, the tool runs structural pre-checks on the raw code; any failure is returned as a `Safeguard rejected` response, and the file is never sent to Lean (so do not assume the code compiles). The pre-checks reject: unbalanced `/- ... -/` block comments; a missing main theorem; forbidden constructs (`axiom`, `native_decide`); missing `import Mathlib` or `import Architect`; a main theorem signature that does not match the targeted signature verbatim (modulo whitespace); a Lemma or Theorem without an `@[blueprint]` attribute; a Lemma/Theorem body that is bare `sorry` or a real proof -- every body must be exactly `:= by sorry_using [...]`, since proofs belong to the next stage and bare `sorry` breaks dependency tracking.

If the pre-checks pass, the code is compiled by Lean. After Lean returns no errors, a post-compile graph-validity check runs against the parsed `@[blueprint]` decls: every node must have a non-empty `(statement := /-- ... -/)` field; every Lemma and the Theorem must have a non-empty `(proof := /-- ... -/)` field; every name in `sorry_using [...]` must resolve to a declared `@[blueprint]` node, with no self-loops; the `sorry_using` graph must be acyclic; exactly one main Theorem must exist with the targeted name; and every node must be reachable, in reverse, from the main Theorem (no isolated/dead nodes).

If any gate fails, fix the reported issue and call `lean_compile` again. Sorries from `sorry_using` are expected and do not count as errors. Iterate until `lean_compile` reports `Compilation SUCCESSFUL. Validation SUCCESSFUL.`
\end{lstlisting}

\subsection{Theorem proving}
\label{app:prompt-blueprint}

\begin{lstlisting}[style=prompt]
## Task
You are a Lean 4 theorem prover. Given a formal statement, produce a complete, correct Lean 4 proof with no `sorry`.

## Tool use
You have two tools, `lean_compile` and `mathlib_search`. Commit to a concrete proof plan up front and execute it against the Lean compiler -- iterating on compiler feedback is how proofs get done, not silent reasoning or repeated searching. The compiler is a stronger signal source than search.

Use `lean_compile` to compile Lean 4 code. Call it early, even with a partial proof: use `sorry` as a placeholder for sub-goals you cannot yet discharge, and iterate (compile -> read errors / open goals -> patch -> compile). The system handles two cases automatically based on what you submit:
- If your code includes the MAIN theorem with the canonical statement followed by `:= by ...`, the system rebuilds under the original theorem statement: only your `:= by` proof body is kept from your submission; the imports, `set_option`, and `open` lines come from the canonical formal statement, and any other top-level declarations are dropped. Only this case can register a solve. Do not use `axiom` or `native_decide`; use `have` for helper lemmas inside your proof, not top-level declarations; and do not add `import` or `open` lines that are not already in the canonical formal statement -- any extras will be flagged as a safeguard violation, not silently kept.
- If your code does NOT include the main theorem (e.g. `#check`, `example`, `#print`, helper-lemma prototypes), the system compiles the snippet as-given and returns the raw feedback. This is exploration only -- it cannot register a solve, so resubmit with the main theorem once you have a full proof. Use this sparingly: every turn against the compiler costs budget, and the only way to finish is to submit the main theorem.

Use `mathlib_search` as a lookup helper for *specific* Mathlib lemmas you need while executing your plan -- for example a name, signature, or hypothesis pattern like "monotonicity of natural number addition" or "Cauchy-Schwarz inequality", or to recover the correct name after an "Unknown constant" / "Unknown identifier" error. Mathlib does NOT contain the solution to your problem directly, so do not use this tool to "find the proof" or to search for an exact bound stated in the goal -- such queries return nothing useful and waste turns.
\end{lstlisting}

\subsection{Blueprint refinement}
\label{app:prompt-reviser}

\begin{lstlisting}[style=prompt]
## Task
You are revising a Lean 4 dependency graph for a single mathematical problem. The input is a sequence of `@[blueprint ...]`-annotated declarations -- definitions, lemmas, and one main theorem -- each lemma or theorem with body `:= by sorry_using [deps]`. Your job is to emit a revised dependency graph -- again all `sorry_using` declarations -- that, when handed back to the same Lean 4 theorem prover, is more likely to close the previously-unsolved nodes while still proving the same main theorem.

## Input format
Each lemma or theorem in the input carries a one-line marker recording the previous prover pass's verdict on that node, and -- when the prover failed -- a follow-up review block describing what went wrong. There are two markers.

A `-- PROVED` marker means the prover proved the node.

A `-- UNPROVED` marker indicates that the prover failed on the node, and is followed by exactly one `/- Diagnosis ... -/` review block. The block has three sections. `## Diagnosis` is exactly one of `STATEMENT_WRONG` (the lemma is false under its hypotheses) or `PROOF_TOO_HARD` (the prover believes the goal is provable but could not chain the available parents to it). `## Analysis` is a forensic account of what the prover tried, what compiled, what errors remained, and where the gap is. `## Suggested Fix` is conditional on the diagnosis: for `STATEMENT_WRONG`, why the statement is false and how to repair it; for `PROOF_TOO_HARD`, a helper-lemma decomposition.

These markers and review blocks are input-only -- do NOT copy them into your revised dependency graph.

## Guidance
Each `-- UNPROVED` node falls into one of two buckets, decided by the `## Diagnosis` label.

When the diagnosis is `STATEMENT_WRONG`, the lemma's formal statement is false under its hypotheses. Fix the statement (strengthen hypotheses, weaken the conclusion, fix a quantifier or coercion, etc.) and re-emit it. If the lemma is structurally unfixable, drop it and re-route the nodes that depended on it.

When the diagnosis is `PROOF_TOO_HARD`, the prover believes the goal is provable but could not chain the available parents to it. Read the `## Suggested Fix` for the prover's proposed helper-lemma decomposition and add new parent lemmas (each as a fresh `@[blueprint ...]` declaration with body `:= by sorry_using [...]`) that bridge the gap. Wire the failing node's `sorry_using [...]` to include the new helpers. If the analysis instead reads as though the statement itself is suspect, treat it as `STATEMENT_WRONG` instead -- fix or drop the statement.

Leave `-- PROVED` nodes untouched unless a downstream revision forces a signature change: their proof bodies will carry forward automatically as long as the signature stays byte-identical.

After every edit, call `lean_compile`. The tool reports pre-compile safeguard violations, real Lean compile errors, the skeleton-out invariant (every theorem/lemma body must remain `:= by sorry_using [...]`), graph-validity issues (cycles, missing fields, dead nodes, etc.), and on a clean compile a per-declaration proof-reuse check. Iterate until `lean_compile` reports `Compilation SUCCESSFUL. Validation SUCCESSFUL.`

## Output
Emit a revised dependency graph. Every theorem and lemma is `@[blueprint (statement := /-- ... -/) (proof := /-- ... -/)]`-annotated and ends in `:= by sorry_using [deps]`. Definitions are `@[blueprint (statement := /-- ... -/)]`-annotated with a real Lean body. Do NOT replace any `sorry_using` with an actual proof -- that is the prover's job, not yours. Preserve the main theorem's signature (name, binders, conclusion) byte-for-byte from the input.
\end{lstlisting}

\end{document}